\documentclass[11pt,a4paper]{article}
\usepackage{authblk}
\usepackage{naaclhlt2019}
\usepackage{times}
\usepackage{latexsym}
\usepackage{array, booktabs, adjustbox, multirow, amssymb}
\usepackage{blindtext}

\usepackage{url}

\aclfinalcopy 
\def\aclpaperid{1499} 

\makeatletter
\newcommand\email[2][]%
   {\newaffiltrue\let\AB@blk@and\AB@pand
      \if\relax#1\relax\def\AB@note{\AB@thenote}\else\def\AB@note{\relax}%
        \setcounter{Maxaffil}{0}\fi
      \begingroup
        \let\protect\@unexpandable@protect
        \def\thanks{\protect\thanks}\def\footnote{\protect\footnote}%
        \@temptokena=\expandafter{\AB@authors}%
        {\def\\{\protect\\\protect\Affilfont}\xdef\AB@temp{#2}}%
         \xdef\AB@authors{\the\@temptokena\AB@las\AB@au@str
         \protect\\[\affilsep]\protect\Affilfont\AB@temp}%
         \gdef\AB@las{}\gdef\AB@au@str{}%
        {\def\\{, \ignorespaces}\xdef\AB@temp{#2}}%
        \@temptokena=\expandafter{\AB@affillist}%
        \xdef\AB@affillist{\the\@temptokena \AB@affilsep
          \AB@affilnote{}\protect\Affilfont\AB@temp}%
      \endgroup
       \let\AB@affilsep\AB@affilsepx
}
\makeatother

\title{Issue Framing in Online Discussion Fora}

\author{\textbf{Mareike Hartmann$^1$ \ \   Tallulah Jansen$^2$ \ \ Isabelle Augenstein$^1$ \ \ Anders S{\o}gaard}}
\affil[$^1$]{Dep. of Computer Science, University of Copenhagen, Denmark}
\email{\tt{\{hartmann,augenstein,soegaard\}@di.ku.dk}}
\affil[$^2$]{Inst. of Cognitive Science, Osnabr{\"u}ck University, Germany}
\email{\tt{taljansen@uni-osnabrueck.de}}

\date{}

\begin{document}
\maketitle
\begin{abstract}
In online discussion fora, speakers often make arguments for or against something, say birth control, by highlighting certain aspects of the topic. In social science, this is referred to as {\em issue framing}. In this paper, we introduce a new issue frame annotated corpus of online discussions. We explore to what extent models trained to detect issue frames in newswire and social media can be transferred to the domain of discussion fora, using a combination of multi-task and adversarial training, assuming only unlabeled training data in the target domain.
\end{abstract}

\section{Introduction}

The \textit{framing} of an issue refers to a choice of perspective, often motivated by an attempt to influence its perception and interpretation \cite{Entman1993,chong2007}. 
The way issues are framed can change the evolution of policy as well as public opinion \cite{Dardis:ea:08, Iyengar1991}. As an illustration, contrast the statement {\em Illegal workers depress wages} with {\em This country is abusing and terrorizing undocumented immigrant workers}. The first statement puts focus on the economic consequences of immigration, whereas the second one evokes a morality frame by pointing out the inhumane conditions under which immigrants may have to work. Being exposed to primarily one of those perspectives might affect the public’s attitude towards immigration.

Computational methods for frame classification have previously been studied in  news articles \cite{Card2015} and social media posts \cite{johnson2017}. In this work, we introduce a new benchmark dataset, based on a subset of the 15 generic frames in the  \textit{Policy Frames Codebook} by \newcite{Boydstun2014}. We focus on frame classification in \textit{online discussion fora}, which have become crucial platforms for public dialogue on social and political issues. Table~1 shows example annotations, compared to previous annotations for news articles and social media. Dialogue data is substantially different from news articles and social media, 
and we therefore explore ways to transfer information from these domains, using multi-task and adversarial learning, providing non-trivial baselines for future work in this area. 


\begin{table}[t]
\begin{small}
\centering
\begin{tabular}{p{6cm}}
\toprule
Platform: Online discussions \\
\midrule
\textbf{Economic} Frame, Topic: Same sex marriage \\
\texttt{But as we have seen, supporting same-sex marriage saves money.}\\
\addlinespace[5pt]
\textbf{Legality} Frame, Topic: Same sex marriage \\
\texttt{So you admit that it is a right and it is being denied?}\\
\toprule
Platform: News articles \\
\midrule
\textbf{Economic} Frame, Topic: Immigration \\
\texttt{Study Finds That Immigrants Are Central to Long Island Economy} \\
\addlinespace[5pt]
\textbf{Legality} Frame, Topic: Same sex marriage \\
\texttt{Last week, the Iowa Supreme Court granted same-sex couples the right to marry.} \\
\toprule
 Platform: Twitter \\
 \midrule
\textbf{Legality} Frame, Topic: Same sex marriage\\
 \texttt{Congress must fight to ensure LGBT people have the full protection of the law everywhere in America. \#EqualityAct} \\
\bottomrule
\end{tabular}
\caption{\label{tab:Example} Example instances from the datasets described in \S\ref{sec:forumdata} and \ref{sec:adddata}.}
\end{small}
\end{table}

\paragraph{Contributions} We present a new issue-frame annotated dataset that is used to evaluate issue frame classification in online discussion fora. Issue frame classification was previously limited to news and social media. As manual annotation is expensive, we explore ways to overcome the lack of labeled training data in the target domain with multi-task and adversarial learning, leading to improved results in the target domain.\footnote{Code and annotations are available at \texttt{\url{https://github.com/coastalcph/issue\_framing}}.}

\paragraph{Related Work}
Previous work on automatic  frame classification focused on news articles and social media. \citet{Card2016} predict frames in news articles at the document level, using clusters of latent dimensions and word-based features in a logistic regression model. \citet{Ji2017} improve on previous work integrating discourse structure into a recursive neural network. \citet{Naderi17a} use the same resource, but make predictions at the sentence level, using topic models and recurrent neural networks. \citet{johnson2017} predict frames in social media data at the micro-post level, using probabilistic soft logic based on lists of keywords, as well as temporal similarity and network structure.  All the work mentioned above uses the generic frames of \citet{Boydstun2014}'s Policy Frames Codebook. \citet{Baumer2015} predict words perceived as frame-evoking in political news articles with hand-crafted features. \citet{Field2018} analyse how Russian news articles frame the U.S. using a keyword-based cross-lingual projection setup. \citet{Tsur2015} use topic models to analyze issue ownership and framing in public statements released by the US congress. Besides work on frame classification, there has recently been a lot of work on aspects closely related to framing, such as subjectivity detection \cite{lin-he-everson:2011:IJCNLP-2011}, detection of biased language \cite{Recasens2013} and stance detection \cite{mohammad-EtAl:2016:SemEval,Augenstein2016stance,FerreiraV16}.

\begin{table}[t]
\centering
\resizebox{0.4\textwidth}{!}{
\begin{tabular}{l ccccc}
\toprule[1pt]
Frames&\texttt{1}& \texttt{13} &\texttt{5} &\texttt{6} &\texttt{7} \\
\# instances &78&96&234&166&186\\
\bottomrule
\end{tabular}
}
\caption{Class distribution in the online discussion test set. The frame labels correspond to the classes \textit{Economic} (\texttt{1}), \textit{Political} (\texttt{13}), \textit{Legality, Jurisprudence and Constitutionality} (\texttt{5}), \textit{Policy prescription and evaluation} (\texttt{6}) and \textit{Crime and Punishment} (\texttt{7}).}\label{t:stats}
\end{table}

\begin{table*}[ht]
\centering
\resizebox{\textwidth}{!}{%
\begin{tabular}{ll l l l r}
\toprule[1pt]
Model & Task & Domain  & Labelset &\# classes & \# sequences \\ 
\midrule
\multirow{2}{*}{Baseline}& \ \ Main task & News articles& Frames &5 &10,480  \\
&\ \ Target task & Online disc. (test)& Frames & 5 & 692\\
\addlinespace[5pt]
\multirow{2}{*}{Multitask}& +Aux task & Tweets & Frames & 5& 1,636 \\
&+Aux task & Online disc. & Argument quality & 2& 3,785\\
\addlinespace[5pt]
Adversarial &+Adv task &  Online disc. + News articles& Domain  & 2 & 4,731 + 10,480\\
\midrule
 && Online disc. (dev)& Frames & 5 & 176\\
\bottomrule
\end{tabular}
}
\caption{Overview over the data and labelsets for the different tasks. The baseline model trains on the main task and predicts the target task. The multi-task model uses one or both auxiliary tasks in addition to the main task. The adversarial model uses the adversarial task in addition to the main task. All models use the online disc. dev set for model selection.}\label{t:tasks}
\end{table*}

\section{Online Discussion Annotations}\label{sec:forumdata}
We create a new resource of issue-frame annotated online fora discussions, by annotating a subset of the Argument Extraction Corpus \cite{Swanson2015} with a subset of the frames in the Policy Frames Codebook. The Argument Extraction Corpus is a collection of argumentative dialogues across topics and platforms.\footnote{The corpus is a combination of dialogues from \url{http://www.createdebate.com/}, and \citet{Walker2012}'s Internet Argument Corpus, which contains dialogues from \url{4forums.com}.} The corpus contains posts on the following topics: \textit{gay marriage}, \textit{gun control}, \textit{death penalty} and \textit{evolution}. A subset of the corpus was annotated with argument quality scores by \citet{Swanson2015}, which we exploit in our multi-task setup (see \S\ref{sec:adddata}).

We collect new issue frame annotations for each argument in the argument-quality annotated data.\footnote{Topic cluster \textit{Evolution} was dropped, because it contained too few examples matching our frame categories.} We refer to this new issue-frame annotated corpus as \textit{online discussion corpus} henceforth. Each argument can have one or multiple frames. Following \citet{Naderi17a}, we focus on the five most frequent issue frames: \textit{Economic}, \textit{constitutionality and jurisprudence}, \textit{policy prescription and evaluation}, \textit{law and order/crime and justice}, and \textit{political}. See Table \ref{tab:Example}~for examples and Table \ref{t:stats}~for the class distribution in the resulting online discussions test set. Phrases which do not match the five categories are labeled as \textit{Other}, but we do not consider this class in our experiments. The annotations were done by a single annotator. A second annotator labeled a subset of 200 instances that we use to compute agreement as macro-averaged F-score, assuming one of the annotations as gold standard. Results are $0.73$ and $0.7$, respectively. The averaged Cohen's Kappa is $0.71$.

\section{Additional Data}\label{sec:adddata}
The dataset described in the previous section serves as evaluation set for the online discussions domain. As we do not have labeled training data for this domain, we exploit additional corpora and additional annotations, which are described in the next subsection. Statistics of the filtered datasets as well as preprocessing details are given in Appendix \ref{sec:appendix:data}. 

\paragraph{Media Frames Corpus}
The Media Frames Corpus \cite{Card2015} contains US newspaper articles on three topics:  \textit{Immigration}, \textit{smoking} and \textit{same-sex marriage}. The articles are annotated with the 15 framing dimensions defined in the Policy Frames Codebook.\footnote{We discard all instances that do not correspond to the frame categories in the online discussions data.} 
The annotations are on span-level and can cross sentence boundaries. We convert span annotations to sentence-level annotations 
as follows: if a span annotated with label $l$ lies within sentence boundaries and covers at least 50\% of the tokens in the sentence, we label the sentence with $l$. We only keep sentence annotations if they are indicated by at least two annotators.

\paragraph{Congressional Tweets Dataset}
The congressional tweets dataset \cite{johnson2017} contains tweets authored by 40 members of the US Congress, annotated with the frames of the Policy Frames Codebook. The tweets are related to one or two of the following six issues: \textit{abortion}, \textit{the Affordable Care Act}, \textit{gun rights vs.~gun control}, \textit{immigration}, \textit{terrorism}, and \textit{the LGBTQ community}, where each tweet is annotated with one or multiple frames.

\paragraph{Argument Quality Annotations}
The corpus of online discussions contains additional annotations that we exploit in the multi-task setup. \citet{Swanson2015} sampled a subset of 5,374 sentences, using various filtering methods to increase likelihood of high quality argument occurrence, and collected annotations for argument quality via crowdsourcing. Annotators were asked to rate argument quality using a continuous slider [0-1]. Seven annotations per sentence were collected. We convert these annotations into binary labels (1 if $\geq$ 0.5, 0, otherwise) and generate an approximately balanced dataset for a binary classification task that is then used as an auxiliary task in the multi-task setup. Balancing is motivated by the observation that balanced datasets tend to be better auxiliary tasks \cite{Bingel:ea:17}. 

\section{Models}
The task we are faced with is (multi-label) sequence classification for online discussions. However, we have no labeled training data (and only a small labeled validation set) for the target task in the target domain. Hence, we train our model on a dataset which is labeled with the target labels, but from a different domain. The largest such dataset is the news articles corpus, which we consequently use as main task. Our baseline model is a two-layer LSTM \cite{Schmidhuber} trained on only the news articles data. We then apply two strategies to facilitate the transfer of information from source to target domain, multi-task learning and adversarial learning. 
We briefly describe both setups in the following. An overview over tasks and data used in the different models is shown in Table \ref{t:tasks}.

\begin{figure}[tb]
\resizebox{.5\textwidth}{!}{
\centering
\includegraphics[]{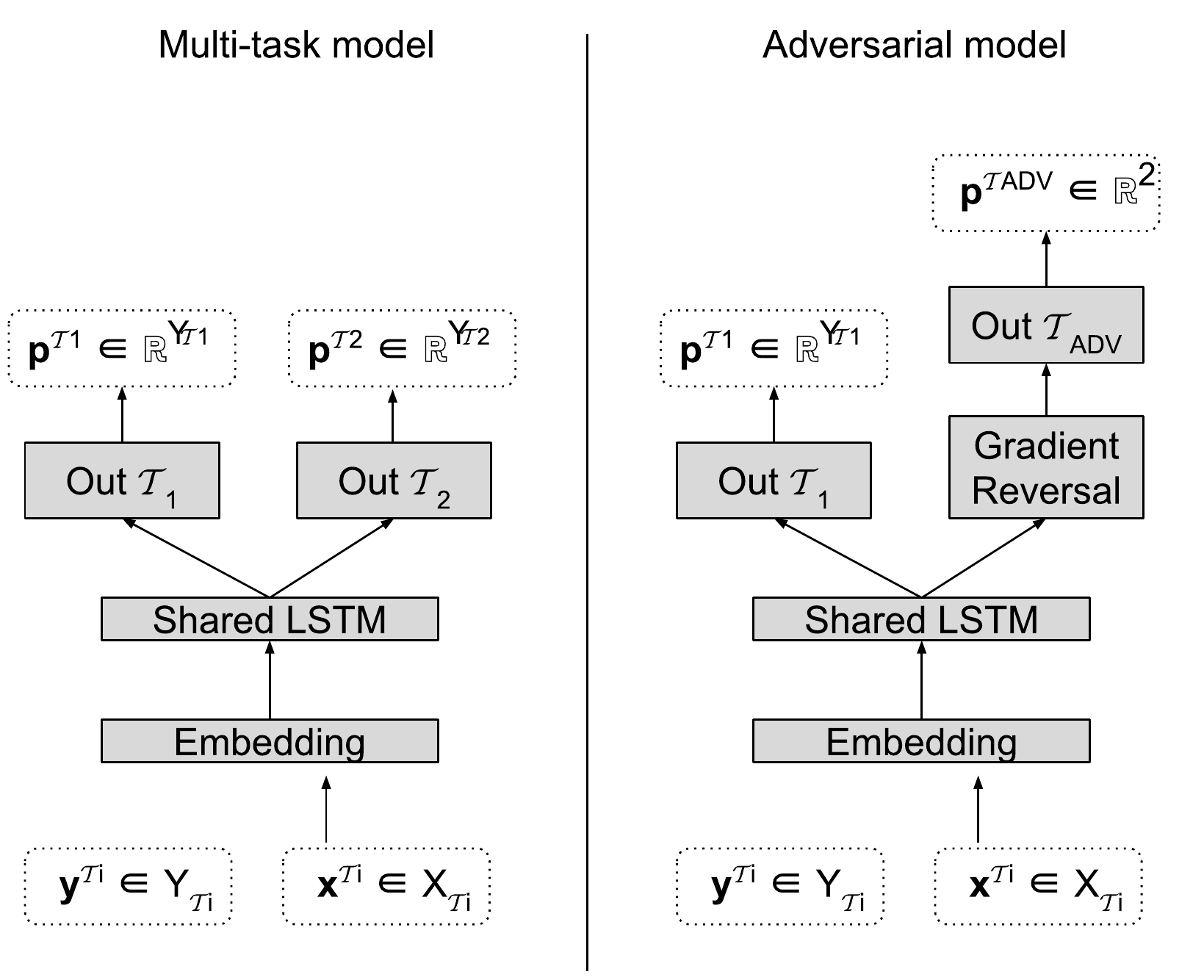}}
\caption{Overview over the multi-task model (left) and the adversarial model (right). The baseline LSTM model corresponds to the same architecture with only one task.
}\label{f:architecture}
\end{figure}

\paragraph{Multi-Task Learning}
To exploit synergies between additional datasets/annotations, we explore a simple multi-task learning with hard parameter sharing strategy, pioneered by \newcite{Caruana93}, introduced in the context of NLP by \newcite{Collobert:ea:11}, and to RNNs by \newcite{Soegaard:Goldberg:16}, which has been shown to be useful for a variety of NLP tasks, e.g. sequence labelling \cite{rei2017ACL,ruder2019aaai,Augenstein2017KBC}, pairwise sequence classification \cite{conf/naacl/AugensteinRS18} or machine translation \cite{dong-EtAl:2015:ACL-IJCNLP2}. Here, parameters are shared between hidden layers. Intuitively, it works by training several networks in parallel, tying a subset of the hidden parameters so that updates in one network affect the parameters of the others. By sharing parameters, the networks regularize each other, and the network for one task can benefit from representations induced for the others. 

Our multi-task architecture is shown in Figure \ref{f:architecture}. We have $N$ different datasets $\mathcal{T}_{1}, \cdots, \mathcal{T}_{N}$. Each dataset $\mathcal{T}_{i}$ consists of tuples of sequences  $x^{\mathcal{T}_i} \in X_{\mathcal{T}_i}$ and labels $y^{\mathcal{T}_i} \in Y_{\mathcal{T}_i}$. A model for task $\mathcal{T}_i$ consists of an input layer, an LSTM layer (that is shared with all other tasks) and a feed forward layer with a softmax activation as output layer. The input layer embeds a sequence $x^{\mathcal{T}_i}$ using pretrained word embeddings. The LSTM layer recurrently processes the embedded sequence and outputs the final hidden state $h$. The output layer outputs a vector of probabilities $p^{\mathcal{T}_i} \in \mathbb{R}^{Y_{\mathcal{T}_i}}$, based on which the loss $\mathcal{L}_i$ is computed as the categorical cross-entropy between prediction $p^{\mathcal{T}_i}$ and true label $y^{\mathcal{T}_i}$. In each iteration, we sample a data batch for one of the tasks and update the model parameters using stochastic gradient descent. If we sample a batch from the main task or an auxiliary task is decided by a weighted coin flip. 

\begin{table*}[t]
\begin{small}
\resizebox{\textwidth}{!}{
\begin{tabular}{ccccc p{12cm}}
\toprule
Nr. & Gold & Adv & MTL & LSTM & Sentence\\
\midrule
(1) & 5 & 5 & 5 & 7 & But, star gazer, we had guns then when the Constitution was written and enshrined in the BOR and now incorporated into th 14th Civil Rights Amendment.\\
\addlinespace[5pt]
(2) & 6 & 6 & 5 & 1 & Gun control is about preventing such security risks.\\
\addlinespace[5pt]
(3) & 7 & 7 & 5 & 1 & First, you warn me of the dangers of using violent means to stop a crime .\\
\addlinespace[5pt]
(4) & 5 & 6 & 6 & 6 & So I don't see restrictions on handguns in D.C. as being a clear violation of the Second Amendment.\\
\bottomrule
\end{tabular}}
\caption{\label{tab:erroranalysis} Examples for model predictions on the online discussion dev set. The first column shows the gold label and the following columns the prediction made by the adversarial model (Adv), the Multi-Task model (MTL) and the LSTM baseline (LSTM).}
\end{small}
\end{table*}

\paragraph{Adversarial Learning}
\citet{Ganin2015} proposed adversarial learning for domain adaptation that can exploit unlabeled data from the target domain. The idea is to learn a classifier that is as good as possible at assigning the target labels (learned on the source domain), but as poor as possible in discriminating between instances of the source domain and the target domain. With this strategy, the classifier learns representations that contain information about the target class but abstract away from domain-specific features. During training, the model alternates between 1) predicting the target labels and 2) predicting a binary label discriminating between source and target instances. In this second step, the gradient that is backpropagated is flipped by a Gradient-Reversal layer.\footnote{In the forward pass, this layer multiplies its input with the identity matrix.} Consequently, the model parameters are updated such that the classifier becomes worse at solving the task. The architecture is shown in the right part of Figure \ref{f:architecture}. In our implementation, the model samples batches from the adversarial task or the main task based on a weighted coinflip.

\section{Experiments}
We compare the multi-task learning and the adversarial setup with two baseline models: (a) a Random Forest classifier using tf-idf weighted bag-of-words-representations, and (b) the LSTM baseline model. For the multi-task model, we use both the Twitter dataset and the argument quality dataset as auxiliary tasks. For all models, we report results on the test set using the optimal hyper-parameters that we found averaged over 3 runs on the validation set. For the neural models, we use 100-dimensional GloVe embeddings \cite{pennington2014glove}, pre-trained on Wikipedia and Gigaword.\footnote{\url{https://nlp.stanford.edu/projects/glove/}} Details about hyper-parameter tuning and optimal settings can be found in Appendix \ref{sec:appendix:hyper}.

\begin{table}[ht]
\resizebox{0.5\textwidth}{!}{%
\begin{tabular}{l c c c c}
\toprule
Model & P$_{ma}$ & R$_{ma}$ & F$_{ma}$ & F$_{mi}$\\
\midrule
Random Baseline & 0.196 & 0.198&0.189 &0.196 \\
\midrule
Random Forest Baseline &0.496 & 0.335& 0.267& 0.279\\
LSTM Baseline & 0.512& 0.510& 0.503&0.521\\
\midrule
Multi-Task &  0.526& 0.525&0.505&0.534\\
Adversarial & \textbf{0.533}&\textbf{0.534}& \textbf{0.515}&\textbf{0.548}
\\
\bottomrule
\end{tabular}
}
\caption{Macro- ($ma$) and micro-averaged ($mi$) scores for the online discussion test data averaged over 3 runs. The multi-task model uses the Twitter and argument quality datasets as auxiliary tasks. The micro-average F of a baseline that predicts the majority class is 0.307.}
\label{t:results}
\end{table}

\begin{figure}[tb]
\resizebox{0.5\textwidth}{!}{
\centering
\includegraphics[width=0.5\textwidth]{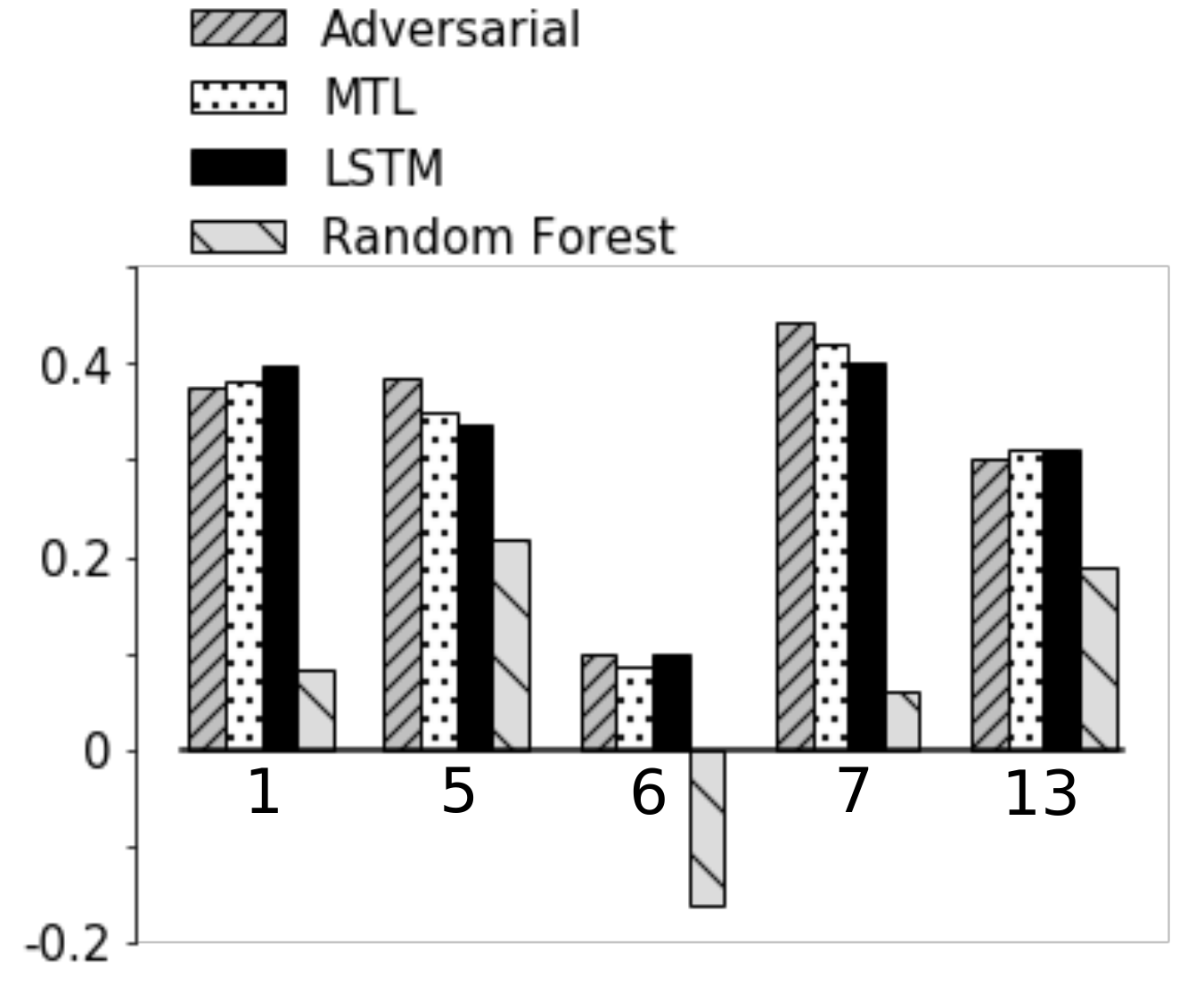}}\caption{Improvement in F-score over the random baseline by class. The absolute F-scores for the best performing system for classes 1, 5, 6, 7, and 13, are 0.529, 0.625, 0.298, 0.655, and 0.499, respectively.}\label{fig:perclass}
\end{figure}

\paragraph{Results}
The results in Table \ref{t:results} show that both the multi-task and the adversarial model improve over the baselines. The multi-task model achieves minor improvements over the LSTM baseline, with a bigger improvement in the micro-averaged score, indicating bigger improvements with frequent labels. The adversarial model performs best, with an error reduction in micro-averaged F over the LSTM baseline of 5.6\%. 

Figure \ref{fig:perclass} shows the system performances for each class. Each bar indicates the difference between the F-score of the respective system and the random baseline. The adversarial model achieves the biggest improvements over the baseline for classes 5 and 7, which are the two most frequent classes in the test set (cf. Table \ref{t:appendix:stats}). For classes 1 and 13, the adversarial model is outperformed by the LSTM. Furthermore, we see that the hardest frame to predict is the \textit{Policy prescription and evaluation frame} (6), where the models achieve the lowest improvement over the baseline and the lowest absolute F-score. This might be because utterances with this frame tend to address specific policies that vary according to topic and domain of the data, and are thus hard to generalize from source to target domain.

\paragraph{Analysis}
Table \ref{tab:erroranalysis} contains examples of model predictions on the dialogue dev set. In Example (1), the adversarial and the multi-task model correctly predict a \textit{Constitutionality} frame, while the LSTM model incorrectly predicts a \textit{Crime and punishment} frame. In Examples (2) and (3), only the adversarial model predicts the correct frames. In both cases, the LSTM model incorrectly predicts an \textit{Economic} frame, possibly because it is misled by picking up on a different sense of the terms \textit{means} and \textit{risks}. In Example (4), all models make an incorrect prediction. We speculate this might be because the models pick up on the phrase \textit{restrictions on handguns} and interpret it as referring to a policy, whereas to correctly label the sentence they would have to pick up on the \textit{violation of the Second Amendment}, indicating a \textit{Constitutionality} frame.

\section{Conclusion}
This work introduced a new benchmark of political discussions from online fora, annotated with issue frames following the Policy Frames Cookbook. Online fora are influential platforms that can have impact on public opinion, but the language used in such fora is very different from newswire and other social media. We showed, however, how multi-task and adversarial learning can facilitate transfer learning from such domains, leveraging previously annotated resources to improve predictions on informal, multi-party discussions. Our best model obtained a micro-averaged F1-score of 0.548 on our new benchmark.  
 
 \section*{Acknowledgements}
We acknowledge the resources provided by CSC in Helsinki through NeIC-NLPL (www.nlpl.eu), and the support of the Carlsberg Foundation and the NVIDIA Corporation with the donation of the Titan Xp GPU used for this research. 
\bibliography{naaclhlt2019}
\bibliographystyle{acl_natbib}

\appendix

\begin{table}[b]
\centering
\resizebox{0.5\textwidth}{!}{
\begin{tabular}{l r | r r r r r| r}
\toprule[1pt]
Dataset & \# instances  & \multicolumn{5}{|c|}{\# instances per class} & \# multi\\ 
\midrule
&&\texttt{1}& \texttt{13} &\texttt{5} &\texttt{6} &\texttt{7} &\\
\midrule
\textsc{Newspaper (Train)} &10,480 &1088&1959&2023&924&890&45\\
\textsc{Twitter (Train)} & 1,636 &73&300&137&27&174&554\\
\textsc{Online Disc. (Test)} &692 &78&96&234&166&186&67\\
\midrule
& & & \texttt{0} & \texttt{1} & & &\\
\midrule
\textsc{Argument Quality} & 3,785 &&1,350& 2,435 & &&0 \\
\midrule
\textsc{Online Disc. unlabeled} & 4731& & &  & & & \\
\addlinespace[5pt]
\end{tabular}
}
\caption{Dataset statistics and class distributions. The frame labels correspond to the classes \textit{Economic} (\texttt{1}), \textit{Political} (\texttt{13}), \textit{Legality, Jurisprudence and Constitutionality} (\texttt{5}), \textit{Policy prescription and evaluation} (\texttt{6}) and \textit{Crime and Punishment} (\texttt{7}). \textit{\# multi} refers to the number of multi-label instances. For Argument quality, label \texttt{1} indicates a score greater or equal 0.5.}\label{t:appendix:stats}
\end{table}

\section{Data Preprocessing}\label{sec:appendix:data}
For the Twitter and news articles datasets, we remove all instances that do not correspond to the five target frames. Table \ref{t:appendix:stats} shows the class distributions in the filtered datasets. We tokenize all sequences using spaCy \footnote{\url{https://spacy.io/}}, which we also use for sentence splitting in the news articles dataset. For the Twitter dataset, we follow \citet{johnson2017} in removing URLs and @-mentions.

\section{Hyperparameters in Experiments}\label{sec:appendix:hyper}
The hyperparameters for all neural models were tuned on the online disc. dev set. We report test results for the optimal settings found by averaging over 3 training runs, which we determine by the best macro-averaged F-score and smallest variance between the runs. We set the DyNet weight decay parameter to 1e-7 for all neural models, batch size is 128, and the word embeddings are not updated during training. 

For the multi-task and adversarial model, we do a grid-search over the weight of the coin flip used to decide on sampling from main/aux or main/adversarial task in the range of [0.1,0.2,0.3,0.4,0.5,0.6,0.7,0.8,0.9]. The optimal weight for sampling the main task is 0.5 for the multi-task model and 0.3 for the adversarial task.

All models are trained using early stopping (after at least 80 epochs of training) with a patience of 5 epochs. The number of iterations (updates) per epoch is a hyperparameter, that we set by default as twice the number of data batches for the main task. For a fair coin flip, the models hence see as much data for the main task as for the auxiliary/adversarial task per epoch.

\end{document}